\documentclass[11pt]{article}

\usepackage[final]{acl}
\usepackage{amsmath}

\usepackage{times}
\usepackage{latexsym}

\usepackage[T1]{fontenc}

\usepackage[utf8]{inputenc}
\usepackage{graphicx}
\usepackage[capitalize,nameinlink]{cleveref}
\usepackage{graphicx}
\usepackage{booktabs}

\usepackage{microtype}
\usepackage{xcolor}

\usepackage{inconsolata}

\usepackage{graphicx}
\usepackage{booktabs}
\usepackage{multirow}
\usepackage{xcolor}
\usepackage{colortbl}
\usepackage{pifont}
\usepackage{tikz}
\usepackage{algorithm}
\usepackage{algpseudocode}
\usepackage{fontawesome5}

\definecolor{natureblue}{RGB}{68,119,170}

\newcommand{\appcref}[1]{Appendix~\ref{#1}}

\newcommand{\fancynumber}[1]{%
\raisebox{1pt}{%
  \tikz[baseline=(char.base)]{
    \node[
      shape=circle,
      draw=black,
      fill=natureblue!20,
      inner sep=0pt,
      minimum size=0.8em,
      font=\tiny,
      text=black,
    ](char){#1};%
  }%
}%
}

\title{Dynamic Hub-and-Spoke Memory for Streaming Video Understanding}

\author{
  \textbf{Xinru Jiang\textsuperscript{1}}%
\thanks{Equal contribution.}%
\thanks{Corresponding authors.},
\textbf{Lin Zhao\textsuperscript{1}}%
\footnotemark[1]%
\footnotemark[2],
 \textbf{Xi Xiao\textsuperscript{2}},
 \textbf{Yunbei Zhang\textsuperscript{3}},\\
 \textbf{Janet Wang\textsuperscript{3}},
 \textbf{Chenrui Ma\textsuperscript{4}},
 \textbf{Haolin Li\textsuperscript{1}},
 \textbf{Yanzhi Wang\textsuperscript{1}},
 \textbf{Yifan Gong\textsuperscript{5}},
 \textbf{Octavia Camps\textsuperscript{1}}
\\
{\small
 \textsuperscript{1}Northeastern University,
 \textsuperscript{2}University of Alabama at Birmingham,
 \textsuperscript{3}Tulane University,}
\\
{\small
\textsuperscript{4}University of Virginia,
 \textsuperscript{5}Adobe Research
}\\
 \small{Please direct correspondence to
\texttt{\{jiang.xinru, zhao.lin1\}@northeastern.edu}.}
 \\
 \small{
 \faGithub\ Project page \url{https://oshikaka.github.io/DHSM/}.}
}

\begin{document}
\maketitle

\begin{abstract}
Streaming video understanding requires answering questions at arbitrary times over a continuously growing visual stream.
The central challenge is to compactly remember long-range history while effectively retrieving question-relevant evidence.
We propose \textbf{D}ynamic \textbf{H}ub-and-\textbf{S}poke \textbf{M}emory \textbf{(D-HSM)}, a \textit{training-free} framework that represents distant history as structured textual memory while preserving the recent frames as visual tokens for fine-grained perception.
Specifically, D-HSM turns selected historical video chunks into typed textual observations and stores them in an entity-centered hub-and-spoke memory, with entities as hubs and related evidence as spokes.
When answering a question, D-HSM dynamically retrieves a compact question-aware memory subset, expands it through hub-and-spoke links, and combines it with the recent visual window for frozen-VLM answer prediction.
Extensive experiments on both streaming and long video benchmarks show that D-HSM consistently and substantially improves VLM backbones and outperforms other state-of-the-art online and offline video understanding baselines.
\end{abstract}

\section{Introduction}
\begin{figure}[t]
    \centering
    \includegraphics[width=\columnwidth]{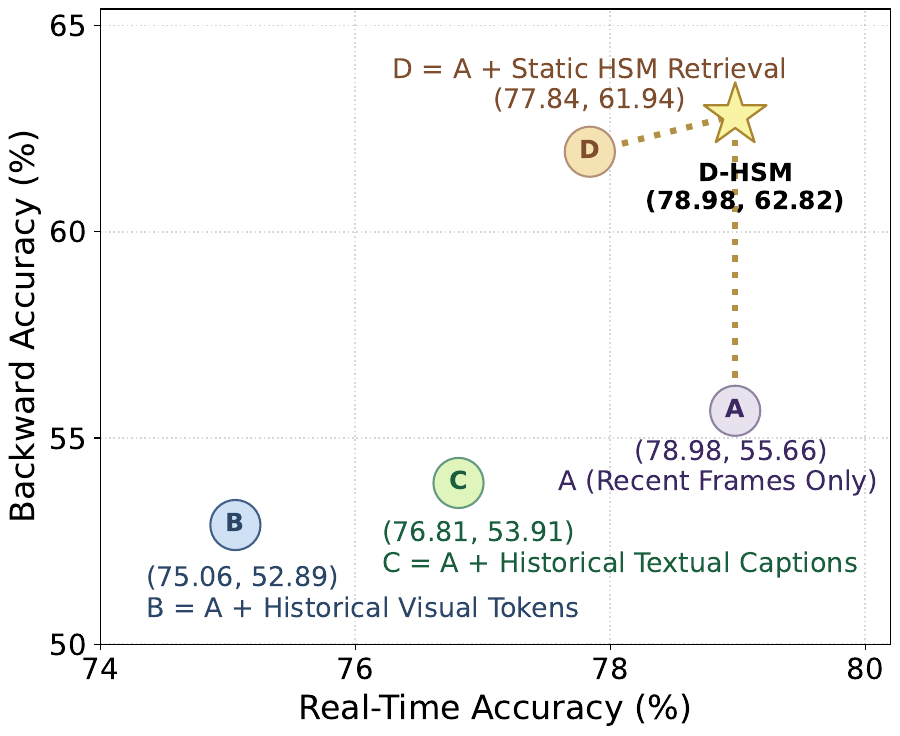}
    \caption{\textbf{Comparison of different historical strategies on OVO-Bench \cite{li2025ovobenchfarvideollmsrealworld}, measured by accuracy on ``Real-Time'' and ``Backward'' tasks.} The ``Static HSM Retrieval'' setting retrieves a fixed top-$K$ budget of memory entries for each question.``Real-Time'' means that the questions can be answered based on the recent video frames, while `Backward'' questions require evidence from history video chunks.}
    \label{fig:motivation}
\end{figure}

\begin{quote}
\textit{``Remembering is not the re-excitation of innumerable fixed, lifeless and fragmentary traces. It is an imaginative reconstruction, or construction.''}\par
\hfill --- Frederic C. Bartlett
\end{quote}

Streaming video understanding faces a memory problem in Bartlett's sense: the model must reconstruct question-relevant evidence rather than merely replay
the past.  
The requirement arises because a vision-language model (VLM) needs
access to long-range visual evidence, while the continuously growing video
history makes it infeasible to preserve the entire stream in raw visual form.
This raises a central question: \textit{how can a streaming model compactly remember the past while efficiently retrieving the evidence?}

Existing streaming video methods largely approach history from a token-budget perspective, reducing the past through sparse token selection~\citep{yao2025timechat,shu2025video,wu2019adaframe}, visual-token compression~\citep{wu2024longvideobenchbenchmarklongcontextinterleaved,li2025videoscanenablingefficientstreaming}, bounded memory banks~\citep{zhang2024flashvstreammemorybasedrealtimeunderstanding,wang2026streammecolongtermagentmemory}, or retrieval over cached visual representations~\citep{di2025streamingvideoquestionansweringincontext,yang2025streammemqueryagnostickvcache}. 
While these mechanisms improve online efficiency, they mainly ask \emph{which} past visual units should be retained, but leave open \emph{how} the retained history should be represented and organized for future questions.
A simple diagnostic in \cref{fig:motivation} illustrates the representation gap: appending historical video tokens (B) is less effective than converting the same segments into textual captions (C).
This suggests that converting distant history into textual semantic traces can provide a useful way to retain it under a limited context budget.
Yet flat captions remain segment-level summaries and lack explicit links among recurring entities, actions, and relations across time.
Therefore, the following question naturally arises: \fancynumber{1} \textit{How should a streaming model organize long-range history into structured semantic memory?}

Besides, different streaming questions require different evidence.
As shown in \cref{fig:motivation}, questions about the current moment are often better answered from the recent visual window alone (A), while adding historical context (B, C, D) can introduce irrelevant entities or events and distract the VLM.
In contrast, backward-looking or temporal questions require evidence from earlier segments, and cannot be answered reliably without retrieving history.
A useful memory must therefore not only be well-organized, but also be queried adaptively.
It should retrieve little or no history for real-time questions, and recover sufficient evidence for questions that depend on the past.
Thus, along with question \fancynumber{1}, another question needs to be answered for streaming video understanding: \fancynumber{2} \textit{How should a streaming model dynamically retrieve question-relevant history?}

To address both questions, we propose \textbf{\textit{D-HSM}}, a training-free \textbf{D}ynamic \textbf{H}ub-and-\textbf{S}poke \textbf{M}emory framework for streaming video understanding.
D-HSM handles the asymmetry between long history and the current moment by storing distant history as compact textual memory, while keeping the recent window as visual frames for fine-grained perception.
%
%
For long-range history, D-HSM first converts selected historical video chunks into structured textual observations that describe visible entities and their associated evidence.
It then organizes these observations into an \textit{entity-centered hub-and-spoke memory}, where recurring entities serve as \textbf{hubs} and related evidence is attached as \textbf{spokes}.
This design turns flat segment captions into structured memory traces, allowing evidence about the same entity to be accumulated, merged, and localized across time.

When answering a question, D-HSM does not replay history in temporal order. Instead, it dynamically retrieves and constructs the relevant evidence on demand.
It first applies keyword-based gating to determine whether memory retrieval is needed.
When retrieval is triggered, D-HSM selects a memory subset according to the similarity between the question and memory entries.
Notably, rather than using a fixed top-$K$ budget, D-HSM determines the retrieval size by detecting a salient cutoff.
It then expands the selected subset through the hub-and-spoke structure.
The retrieved textual evidence of the subset is then combined with the recent visual tokens and passed to a frozen VLM for answer prediction.
As illustrated by \cref{fig:motivation}, D-HSM reconstructs question-relevant evidence from structured history while avoiding distraction from unnecessary historical information.

Together, the structured hub-and-spoke memory and the question-adaptive retrieval give a frozen VLM a way to remember the past compactly and access it selectively, without any additional training.
Extensive experiments show that D-HSM achieves state-of-the-art performance on streaming benchmarks such as StreamingBench \cite{lin2024streamingbenchassessinggapmllms} and OVO-Bench \cite{li2025ovobenchfarvideollmsrealworld}, outperforming strong proprietary models, as well as open-source online and offline video understanding baselines.
Meanwhile, D-HSM also retains strong performance on conventional offline long-video understanding benchmarks, including LongVideoBench \cite{wu2024longvideobenchbenchmarklongcontextinterleaved}, MLVU \cite{zhou2025mlvubenchmarkingmultitasklong}, and VideoMME \cite{fu2025videommefirstevercomprehensiveevaluation}.
These results demonstrate the robustness and generality of D-HSM for compactly remembering long video history and retrieval question-relevant evidence on demand.

\section{Related Work}
\begin{figure*}[t]
    \centering
    \includegraphics[width=\linewidth]{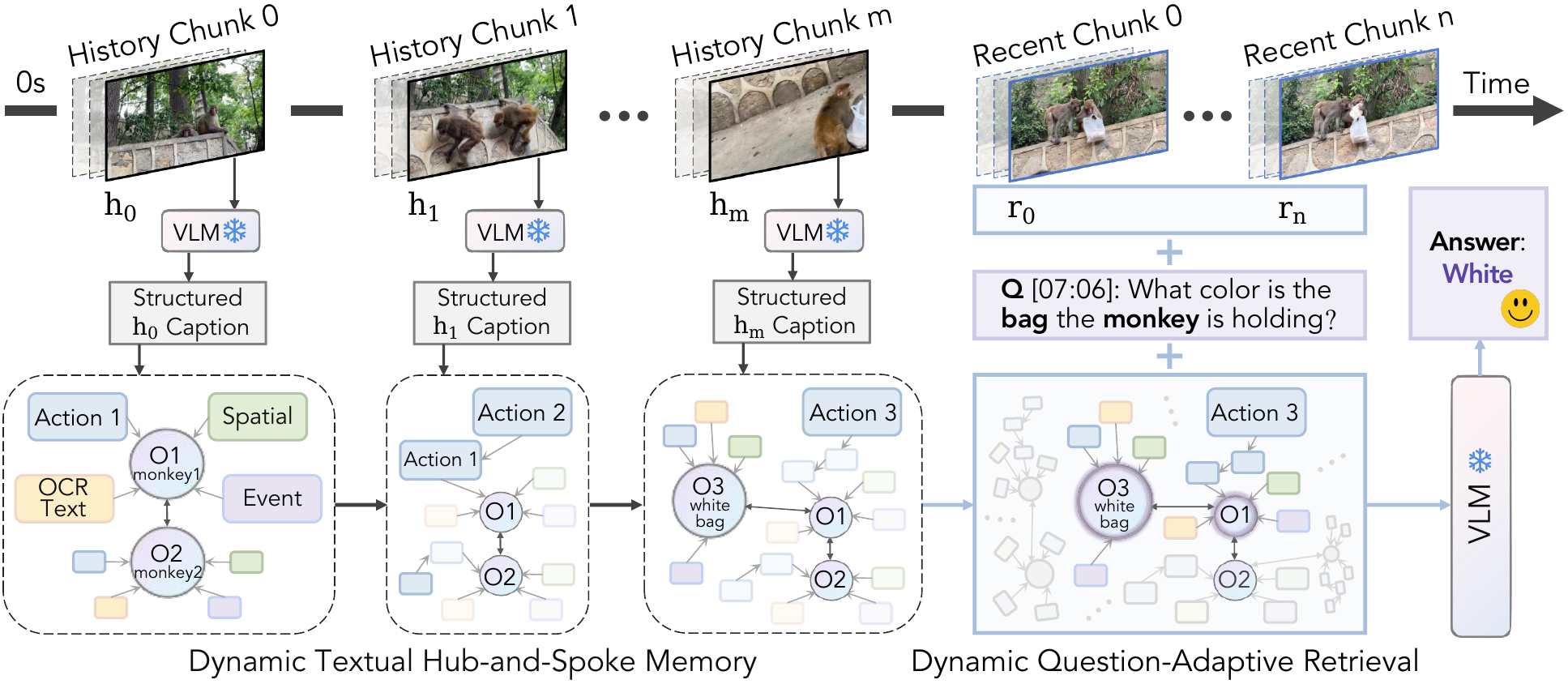}
    \caption{\textbf{Overview of D-HSM.}
Historical chunks are converted by a frozen VLM into structured textual observations and incrementally organized into a dynamic hub-and-spoke memory.
When a question arrives, D-HSM dynamically retrieves a compact question-adaptive subset from memory and expands it through hub-and-spoke links.
Then, it combines the retrieved textual evidence with the recent visual tokens for frozen-VLM answer prediction.}
    \label{fig:framework}
\end{figure*}

\textbf{Long and Streaming Video Understanding.}
Recent methods address long-range temporal reasoning mainly by retaining or compressing visual tokens. Offline long-video models~\citep{chen2025longvila,wang2024longllava,shu2025video,song2024moviechat,wang2024videoagent,corona2025vlogger,lin2026unleashing} extend visual context by sampling more frames or scaling representations, but their computational costs grow rapidly with video length. Some methods organize events into graph memories, retrieving only query relevant subgraph to reduce the cost~\citep{chu2025understanding,huang2026buildingmindpalacestructuring,malik2026ravu}. They are primarily evaluated in offline  settings rather than under streaming queries. 
Online or streaming approaches~\citep{chen2024videollm,huang2024online,zeng2026streamforest,carreira2017quo,feichtenhofer2019slowfast,bertasius2021space,arnab2021vivit,liu2026graph2video,zhangprogressive} instead process frames sequentially. To manage history, sampling-based compression~\citep{yao2025timechat,shu2025video,wu2019adaframe,liu2025keyframe} reduces input tokens but discards fine-grained spatial or temporal details. Alternatively, storage-based compression~\citep{zeng2026streamforest,gu2024mamba,sun2024hawk,wang2025accelerating,li2024videomamba,he2024ma, zheng2026pearl} maintains visual memory after encoding. 
These methods largely retain history as features without explicitly organizing the relationship between different entities.
D-HSM instead maintains a persistent, entity-centric memory that is incrementally updated as chunks arrive, enabling compact and question adaptive retrieval.


\noindent \textbf{Video Large Language Models.}
Video Large Language Models (Video-LLMs) map spatiotemporal visual features into LLM spaces to enable open-ended dialogue and reasoning~\cite{maaz2024video,lin2024video,wang2024internvideo2,li2023blip,liu2024llava,achiam2023gpt,xiao2026staying}. Standard architectures establish modality alignment via abstractors~\cite{li2023blip,ye2023mplug,zhao2024thinimg}, projection layers~\cite{liu2024llava,zhu2023minigpt,li2026rethinking,xiao2026adapting}, or temporal pooling operators~\cite{zhang2023video} over established vision backbones~\cite{radford2021learning,fang2023eva,zhao2026hieramp,zhao2026s2dit}. Recent variants optimize efficiency via dynamic resolution encoding~\cite{bai2025qwen} and parameter-efficient tuning~\cite{hu2021lora,xiao2026not,zhang2025sensitivity}.
%
\section{Method}
\begin{figure*}[t]
    \vspace{-1mm}
    \centering
    \includegraphics[width=0.9\linewidth]{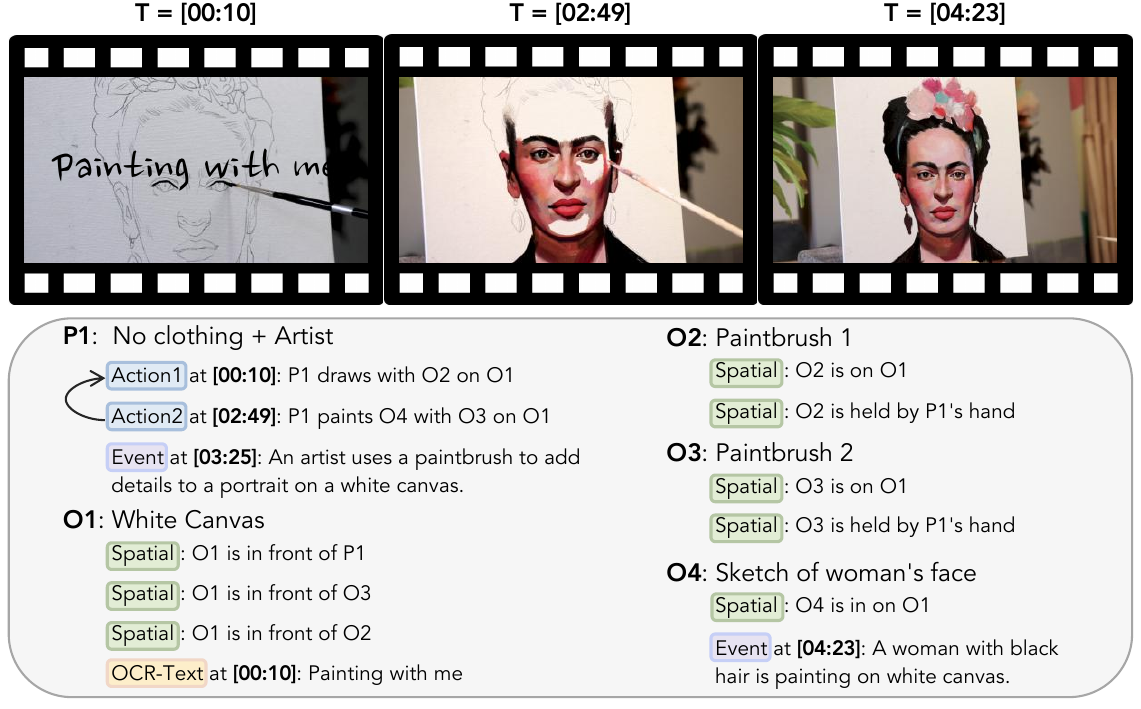}
    \caption{\textbf{Example of the hub-and-spoke memory representation.}
Entities such as the ``artist'', ``canvas'', ``paintbrushes'', and ``sketch'' are stored as hubs with consistent identifiers, while their associated actions, spatial relations, OCR text, and events are stored as timestamped spokes.
Repeated or related evidence is attached to the corresponding entity hub, enabling D-HSM to localize and connect evidence across different moments in the video.}
    \label{fig:hs}
\end{figure*}

\subsection{Framework}
For the streaming video understanding, we require balance two forms of evidence: long-range historical context and immediate perception \cite{zeng2026streamforest, lin2024streamingbenchassessinggapmllms, qian2024streaminglongvideounderstanding}.
These two information streams are inherently asymmetric, as long-range history is too voluminous to retain in raw form, while the present moment demands fine-grained visual detail that resists lossy abstraction.
Building on this observation, we propose Dynamic Hub-and-Spoke Memory (D-HSM), a training-free framework that preserves recent video evidence as visual frames and consolidates longer-range history into a textual hub-and-spoke memory of compact semantic traces.

Specifically, as shown in Figure~\ref{fig:framework}, D-HSM maintains a textual memory $M_t$ for the historical stream.  
At each streaming step $t$, the historical chunk $h_t$ is converted into a typed textual observation $z_t$, which is incrementally integrated into the entity-centered hub-and-spoke memory:
\[
z_t = \phi_{\mathrm{obs}}(h_t), \quad
M_t = \mathcal{U}(M_{t-1}, z_t), 
\]
where $M_t$ is the textual hub-and-spoke memory accumulated up to step $t$.


%
Questions may arrive at arbitrary streaming steps. When a question $q_t$ arrives at step $t$, D-HSM dynamically selects a compact question-aware hub-and-spoke evidence subset $C_t$ from the current memory $M_t$ and combines it with the recent visual window $r_t$:
\[
C_t = \mathcal{R}(M_t, q_t), \quad
a_t = f_{\mathrm{VLM}}(C_t, r_t, q_t),
\]
where $\mathcal{R}$ denotes dynamic hub-and-spoke retrieval, $f_{\mathrm{VLM}}$ is the frozen VLM used for answer prediction, and $a_t$ is the predicted answer.

\subsection{D-HSM Construction and Update}
\label{sec:method-memory}
D-HSM constructs the historical textual memory $M_t$ in three stages: (i) it dynamically selects a bounded set of historical video chunks, (ii) it converts the selected chunks into structured textual observations, and (iii) it integrates these observations into an entity-centered Hub-and-Spoke memory.

\noindent\textbf{Dynamic Historical Observation Budget.}
Before constructing memory, D-HSM adapts the density of historical observations to the length of the available video prefix. 
We construct historical chunks by uniformly sampling frames from the video, with each sampled frame treated as the representative of one chunk.
Let $\mathcal{H}_t=\{h_1,\ldots,h_t\}$ be the candidate historical chunks at step $t$.
D-HSM selects an active observation set:
\[
\mathcal{S}_t =
\begin{cases}
\mathcal{H}_t, & t \le B,\\
\Pi_B(\mathcal{H}_t), & t > B,
\end{cases}
\]
where $B$ is the history budget and $\Pi_B(\cdot)$ denotes approximately uniform selection of $B$ chunks from $\mathcal{H}_t$.
As the video grows, the change in the active set is:
\[
\Delta_t^+ = \mathcal{S}_t \setminus \mathcal{S}_{t-1}, \quad
\Delta_t^- = \mathcal{S}_{t-1} \setminus \mathcal{S}_t .
\]
D-HSM generates textual observations only for chunks in $\Delta_t^+$, and excludes evidence supported only by chunks in $\Delta_t^-$ from the active memory.
This allows short prefixes to be represented densely, while longer prefixes are represented by a sparser set of semantic traces.

\noindent\textbf{Structured Textual Observations.} 
For each selected historical chunk, D-HSM applies a fixed-schema prompt (details in \appcref{app:observation-template}) to instruct a frozen VLM, yielding a textual observation: \textsc{Objects}, \textsc{People}, \textsc{Actions}, \textsc{OCR-Text}, \textsc{Spatial}, and \textsc{Event}.
%
\textsc{Objects} and \textsc{People} produce chunk-level entity mentions with local identifiers and concise visual descriptions. Across different chunks, D-HSM merges mentions of the same entity into a single persistent identity based on the similarity of their visual descriptions.
\textsc{Actions} describes interactions among these entities, \textsc{OCR-Text} captures screen text across optical character recognition, \textsc{Spatial} records positional relations, and \textsc{Event} summarizes the main event in the current chunk.

The schema converts each chunk into typed evidence rather than an unstructured caption.  
Persistent entity identifiers make it easier to track the same person or object across chunks by assigning each visible entity a unique symbol, such as $P_i$ or $O_i$, and reusing the same symbol whenever that entity reappears.
The remaining fields preserve complementary evidence in separable forms.  
Together, these typed observations expose the information needed for later memory construction and retrieval.

\noindent\textbf{Entity Hub-Centered Memory Construction and Update.} 
As shown in \cref{fig:hs}, D-HSM is organized around \emph{entity hubs} instantiated from \textsc{Objects} and \textsc{People}, which provide anchors for \emph{who} or \emph{what} is tracked across time.
%
Other chunk-level evidence types are stored as \emph{spokes} attached to the relevant entity hubs. 
These spokes answer \emph{what happened}, \emph{where it happened}, and \emph{when it was observed}.  
Besides, each hub and spoke also stores its timestamps and occurrence count, allowing D-HSM to localize evidence in time.
%

As illustrated in \cref{fig:framework}, D-HSM updates the Hub-and-Spoke memory online.
For each incoming observation at time $t$, it applies the following rules:
\textit{(i) Entity linking and hub update.}
For each newly observed entity, D-HSM compares its description embedding with the embeddings of existing entity hubs using cosine similarity. If the similarity to the closest hub exceeds $\tau_{\mathrm{link}}$, the entity is merged into that hub and reuses its persistent identifier. Otherwise, a new hub is created. The
resulting hub then records the supporting chunk and timestamp.
\textit{(ii) Spoke insertion and merging.} For spokes that mention entity identifiers such as \textit{P1} or
\textit{O2}, D-HSM attaches the spoke to the corresponding entity hub.
When the spoke text does not mention any entity identifier, D-HSM compares the spoke embedding with existing entity-hub embeddings, and links it to the most similar hub.
Then, D-HSM updates existing spokes for repeated facts by refreshing their timestamps and occurrence count, while inserting novel facts as new spokes.
\textit{(iii) Local relational update.} Entities that appear in the same chunk are connected with co-occurrence edges.  These edges record local entity context, so retrieving one entity can also surface other entities observed in the same chunk.
\textit{(iv) Temporal action update.} For each entity hub, D-HSM links action spokes according to their observation order. These edges record the local order of actions performed by the same entity, so a retrieved action can bring in neighboring actions performed by the same entity.
\textit{(v) Support-based memory removal.}
D-HSM records the supporting chunks for each memory element and
maintains an inverted index from chunks to their contributions.
When a chunk in $\Delta_t^{-}$ is removed, D-HSM deletes its support,
recomputes the affected counts and timestamps, and removes elements
with no remaining support. Spoke attachments and relational edges
are then updated using the remaining observations, preventing stale
evidence from remaining in memory.

\subsection{Dynamic Hub-and-Spoke Retrieval}
\label{sec:method-retrieval}

Given the current memory $M_t$ and query $q_t$, D-HSM retrieves a compact, question-aware subset of the Hub-and-Spoke memory rather than passing the entire memory to the VLM.  
Retrieval proceeds in two steps: query-adaptive evidence selection and hub-and-spoke evidence expansion.

\noindent\textbf{Question-Adaptive Evidence Selection.}
D-HSM first uses a lightweight keyword-based gate to skip memory retrieval for explicitly current-state questions (e.g., cues such as \textit{now}, \textit{currently}, or \textit{recent frame}).
For history-dependent questions, D-HSM embeds the query $q_t$ and the memory entries, and computes cosine similarity scores. 
After applying a base threshold $\theta$ and a maximum budget $K$, we obtain a ranked candidate list
$\{(e_i,s_i)\}_{i=1}^{n}$, where $s_i \ge s_{i+1}$ and $n \le K$.
D-HSM truncates this list at the most prominent \emph{score gap}, when such a gap is statistically distinguishable from the rest of the distribution. Specifically, we examine the gaps:
\[
\Delta_i = s_i - s_{i+1}, \quad i = 1, \ldots, n-1,
\]
and locate the largest one, $\Delta^\star = \Delta_{i^\star}$ with $i^\star = \arg\max_i \Delta_i$.
We accept $i^\star$ as the truncation point only if $\Delta^\star$ passes two tests. First, it must exceed a confidence-adaptive threshold:
\[
\gamma_t = \max\!\big(\lambda_0,\; \lambda_1[s_1 - \theta]_+\big), \quad [x]_+ = \max(x, 0),
\]
where $\lambda_0$ is an absolute floor and the second term tightens the threshold when the top score $s_1$ is high, so that a sharper gap is required when the retriever is confident. Second, $\Delta^\star$ must be at least twice the median gap, $\Delta^\star \ge 2\,\mathrm{median}(\{\Delta_i\})$, which rules out cases where scores decay smoothly and no single gap is genuinely salient.
When both tests pass, we set $k_t = i^\star$; otherwise (including the degenerate case $n < 2$) we fall back to $k_t = n$. D-HSM then retrieves $\mathcal{E}_t = \{e_i\}_{i=1}^{k_t}$.


\begin{table*}[t]
\centering
\normalsize
\caption{\textbf{Comparison results of different methods on StreamingBench  Real-Time
understanding tasks \cite{lin2024streamingbenchassessinggapmllms}.} D-HSM is evaluated with Qwen2.5-VL \cite{bai2025qwen25vltechnicalreport} and Qwen3-VL \cite{bai2025qwen3vltechnicalreport} using $4$ or $8$ recent frames, and compared with different proprietary, open-source offline, and open-source online video understanding models. ``20+4'' and ``20+8'' denote using 20 historical frames for memory construction and 4 or 8 recent frames for visual perception.}
\resizebox{\linewidth}{!}{
\begin{tabular}{llc|cccccccccc|c}
\toprule
\textbf{Model} & \textbf{Size} & \textbf{\#Frames} & \textbf{OP} & \textbf{CR} & \textbf{CS} & \textbf{ATP} & \textbf{EU} & \textbf{TR} & \textbf{PR} & \textbf{SU} & \textbf{ACP} & \textbf{CT} & \textbf{Overall} \\
\midrule
\rowcolor{gray!10}Human  & -  & - & 89.5 & 92.0 & 93.6 & 91.5 & 95.7 & 92.5 & 88.0 & 88.8 & 89.7 & 91.3 & 91.5\\
\midrule
\multicolumn{14}{c}{\textsc{Proprietary Models}} \\

GPT-4o         & -  & 64 & 77.1 & 80.5 & 83.9 & 76.5 & 70.2 & 83.8 & 66.7 & 62.2 & 69.1 & 49.2 & 73.3 \\
Claude 3.5 Sonnet  & -  & 20 & 80.5 & 77.3 & 82.0 & 81.7 & 72.3 & 75.4 & 61.1 & 61.8 & 69.3 & 43.1 & 72.4\\
Gemini 1.5 pro  & -  & 1 fps & 79.0 & 80.5 & 83.5 & 79.7 & \textbf{80.0} & 84.7 & 77.8 & 64.2 & 72.0 & 48.7 & 75.7 \\

\midrule
\multicolumn{14}{c}{\textsc{Open-source Offline Models}} \\

VILA-1.5     & 8B & 14 & 53.7 & 49.2 & 71.0 & 56.9 & 53.4 & 53.9 & 54.6 & 48.8 & 50.1 & 17.6 & 52.3 \\
LongVA       & 7B & 128 & 70.0 & 63.3 & 61.2 & 70.9 & 62.7 & 59.5 & 61.1 & 53.7 & 54.7 & 34.7 & 60.0 \\
MiniCPM-v2.6 & 7B & 32 & 71.9 & 71.1 & 77.9 & 75.8 & 64.6 & 65.7 & 70.4 & 56.1 & 62.3 & 53.4 & 67.4 \\
LLaVA-OneVision & 7B & 32 & 80.4 & 74.2 & 76.0 & 80.7 & 72.7 & 71.7 & 67.6 & 65.5 & 65.7 & 45.1 & 71.1 \\
\rowcolor{gray!10} Qwen2.5-VL    & 7B & 1 fps & 78.3 & 80.5 & 78.9 & 80.5 & 76.7 & 78.5 & 79.6 & 63.4 & 66.2 & 53.2 & 73.7 \\

\midrule
\multicolumn{14}{c}{\textsc{Open-source Online Models}} \\

Flash-VStream   & 7B & 1 fps & 25.9 & 43.6 & 24.9 & 23.9 & 27.3 & 13.1 & 18.5 & 25.2 & 23.9 & 48.7 & 23.2 \\
VideoLLM-online  & 8B & 2 fps & 39.1 & 40.1 & 34.5 & 31.1 & 46.0 & 32.4 & 31.5 & 34.2 & 42.5 & 27.9 & 36.0 \\
Dispider        & 8B & 1 fps & 74.9 & 75.5 & 74.1 & 73.1 & 74.4 & 59.9 & 76.1 & 62.9 & 62.2 & 45.8 & 67.6 \\
TimeChatOnline  & 7B & 1 fps & 80.2 & \underline{82.0} & 79.5 & 83.3 & 76.1 & 78.5 & 78.7 & 64.6 & 69.6 & \textbf{58.0} & 75.4 \\
Streamforest    & 7B & 1 fps & 83.1 & \textbf{82.8} & 82.7 & 84.3 & 77.5 & 78.2 & 76.9 & 69.1 & 75.6 & 54.4 & 77.3 \\

\midrule
\rowcolor{cyan!3}\textit{Qwen2.5-VL}+D-HSM (4f) & 7B  & 20+4 & {85.3} & {64.0} & {88.8} & {87.9} & {79.1} & {90.0} & {\textbf{87.6}} & \underline{79.7} & {79.6} & {47.8} & {82.5} \\
\rowcolor{cyan!7}\textit{Qwen2.5-VL}+D-HSM (8f) & 7B & 20+8 & 88.0 & 70.4 & \underline{93.1} & 88.2 & 78.5 & \textbf{93.8} & 81.9 & \textbf{80.9} & \underline{83.1} & 50.0 & \underline{84.7} \\
\rowcolor{cyan!3}\textit{Qwen3-VL}+D-HSM (4f) & 8B & 20+4 & \underline{86.9} & {64.8} & 92.1 & \underline{89.2} & {74.7} & {90.7} & \underline{85.7} & {77.6} & {79.6} & {54.5} & {83.1} \\
\rowcolor{cyan!7}\textit{Qwen3-VL}+D-HSM (8f) & 8B & 20+8 & \textbf{88.6} & {67.2} & \textbf{95.1} & \textbf{90.8} & \underline{79.8} & \underline{92.5} & \textbf{87.6} & 77.2 & \textbf{84.0} & \underline{55.6} & \textbf{85.4} \\
\bottomrule
\end{tabular}
}

\label{tab:video_results}
\end{table*}

\noindent\textbf{Hub-and-Spoke Evidence Expansion.}
Given the selected evidence entries $\mathcal{E}_t$, D-HSM expands them using the Hub-and-Spoke structure of the memory.  
For an entity hub, the expansion includes its attached spokes, together with entities that co-occurred in the same chunk as discussed in \cref{sec:method-memory}. 
Notably, when attaching action spokes, D-HSM follows a short next-action chain to recover neighboring actions performed by the same entity. For time-aware questions, it also retrieves timestamped events from the stored timeline.

Then, the resulting evidence set is linearized into a textual context with entity-centered profiles, screen-text snippets, chronological event lines, and counting aggregates as illustrated in \cref{tab:rendered-context}. 
This context is then combined with the recent visual frames and the query for frozen-VLM.

\section{Experiments}
\begin{table*}[t!]
\centering
\normalsize
\caption{\textbf{Comparison results of different methods on OVO-Bench \cite{li2025ovobenchfarvideollmsrealworld}.} }
\resizebox{\linewidth}{!}{
\begin{tabular}{lcc|ccccccc|cccc|cccc|c}
\toprule
\multirow{2}{*}{\textbf{Model}} & \multirow{2}{*}{\begin{tabular}[c]{@{}c@{}} \textbf{Size}\end{tabular}} & \multirow{2}{*}{\textbf{\#Frames}} & \multicolumn{7}{c|}{\textbf{Real-Time}} & \multicolumn{4}{c|}{\textbf{Backward}} & \multicolumn{4}{c|}{\textbf{Forward}} & \textbf{Overall} \\ 

\cmidrule(lr){4-10} \cmidrule(lr){11-14} \cmidrule(lr){15-18}
 &  &  & OCR & ACR & ATR & STU & FPD & OJR & \textbf{Avg.} & EPM & ASI & HLD & \textbf{Avg.} & REC & SSR & CRR & \textbf{Avg.} & \textbf{Avg.} \\ 

\midrule
\rowcolor{gray!10}Human Agents & -  & - & 94.0 & 92.6 & 94.8 & 92.7 & 91.1 & 94.0 & 93.2 & 92.6 & 93.0 & 91.4 & 92.3 & 95.5 & 89.7 & 93.6 & 92.9 & 92.8\\
\midrule

\multicolumn{19}{c}{\textsc{Proprietary Models}} \\ 
GPT-4o & - & 64 & 69.8 & 64.2 & 71.6 & 51.1 & 70.3 & 59.8 & 64.5 & 57.9 & \underline{75.7} & 48.7 & 60.8 & 27.6 & \underline{73.2} & 59.4 & 53.4 & 59.5 \\ 
Gemini 1.5 pro & - & 1 fps & 85.9 & 67.0 & 79.3 & 58.4 & 63.4 & 62.0 & 69.3 & \underline{58.6} & \textbf{76.4} & 52.6 & \underline{62.5} & 35.5 & \textbf{74.2} & 61.7 & 57.2 & 63.0 \\
\midrule
\multicolumn{19}{c}{\textsc{Open-source Offline Models}} \\ 
LongVU & 7B & 1 fps & 53.7 & 53.2 & 62.9 & 47.8 & 68.3 & 59.8 & 57.6 & 40.7 & 59.5 & 4.8 & 35.0 & 12.2 & 69.5 & 60.8 & 47.5 & 46.7 \\
LLaVA-OV & 7B & 64 & 66.4 & 57.8 & 73.3 & 53.4 & 71.3 & 62.0 & 64.0 & 54.2 & 55.4 & 21.5 & 43.7 & 25.6 & 67.1 & 58.8 & 50.5 & 52.7 \\
LLaVA-Video & 7B & 64 & 69.1 & 58.7 & 68.8 & 49.4 & \underline{74.3} & 59.8 & 63.5 & 56.2 & 57.4 & 7.5 & 40.4 & 34.1 & 70.0 & 60.4 & 54.8 & 52.9 \\
Qwen2-VL & 72B & 64 & 65.8 & 60.6 & 69.8 & 51.7 & 69.3 & 54.4 & 61.9 & 52.5 & 60.8 & 57.5 & 57.0 & \textbf{38.8} & 64.1 & 45.0 & 49.3 & 56.3 \\
\midrule
\multicolumn{19}{c}{\textsc{Open-source Online Models}} \\ 
VideoLLM-online & 8B & 2 fps & 8.1 & 23.9 & 12.1 & 14.0 & 45.5 & 21.2 & 20.8 & 22.2 & 18.8 & 12.2 & 17.7 & - & - & - & - & - \\
Dispider & 8B & 1 fps & 57.7 & 49.5 & 62.1 & 44.9 & 61.4 & 51.6 & 54.6 & 48.5 & 55.4 & 4.3 & 36.1 & 18.1 & 37.4 & 48.8 & 34.7 & 41.8 \\
TimeChatOnline & 7B & 1 fps & 75.2 & 46.8 & 70.7 & 47.8 & 69.3 & 61.4 & 61.9 & 55.9 & 59.5 & 9.7 & 41.7 & 31.6 & 38.5 & 40.0 & 36.7 & 46.7 \\
Streamforest & 7B & 1 fps & 68.5 & 53.2 & 71.6 & 47.8 & 65.4 & 60.9 & 61.2 & \textbf{58.9} & 64.9 & 32.3 & 52.0 & 32.8 & 70.6 & 57.1 & 52.5 & 55.6 \\
Streamo & 7B & 1 fps & 77.2 & 66.1 & 76.7 & 45.5 & 66.3 & 72.8 & 67.4 & 55.6 & 58.1 & 33.9 & 49.2 & 30.8 & 57.6 & \textbf{82.5} & 57.0 & 57.9 \\

\midrule
\rowcolor{cyan!3}
\textit{Qwen2.5-VL}+ D-HSM (4f) & 7B & 20+4
& 94.0 & 70.7 & \textbf{86.2} & 66.9 & \textbf{75.3} & 81.0 & 79.0 
& 52.9 & 63.5 & \textbf{72.0} & \textbf{62.8} 
& \underline{36.4} & 71.9 & 67.1 & \underline{58.5} & \textbf{66.8} \\ 
\rowcolor{cyan!7}
\textit{Qwen2.5-VL}+ D-HSM (8f) & 7B & 20+8
& \textbf{96.0} & 73.4 & 81.0 & 66.3 & \textbf{75.3} & \underline{82.1} & 79.0 
& 52.2 & 60.8 & 62.4 & 58.5  
& \underline{36.4} & \textbf{74.2} & 67.1 & \textbf{59.2} & \underline{65.6} \\ 
\rowcolor{cyan!3}
\textit{Qwen3-VL}+ D-HSM (4f) & 8B & 20+4
& \underline{95.0} & \textbf{82.6} & \underline{84.5} & \textbf{72.5} & 71.3 & \textbf{83.7} & \textbf{81.6} 
& 54.6 & 62.2 & \underline{67.2} & 61.3
& 25.0 & 61.9 & \underline{74.6} & 53.8 & \underline{65.6} \\ 
\rowcolor{cyan!7}
\textit{Qwen3-VL}+ D-HSM (8f) & 8B & 20+8
& 94.0 & \underline{81.7} & 80.2 & \underline{67.4} & 72.3 & 81.5 & \underline{79.5} 
& 55.2 & 65.5 & 62.4 & 61.1  
& 24.9 & 64.7 & \underline{74.6} & 54.7 & 65.1\\
\bottomrule
\end{tabular}
}

\label{tab:results}
\end{table*}
\subsection{Implementation Details.}
D-HSM is training-free and keeps all VLM parameters frozen.
We use Qwen2.5-VL-7B \cite{bai2025qwen25vltechnicalreport} and Qwen3-VL-8B as backbone models \cite{bai2025qwen3vltechnicalreport}.
For each video, we use $20$ historical chunks by default to construct the hub-and-spoke memory.
%
We set 4 recent frames and  dynamic retrieval with a maximum budget of $K=12$ as default setting.
Memory entries and questions are embedded with a lightweight embedding encoder (bge-small-en-v1.5) \cite{bge_embedding}, and the dynamic cutoff gap $\lambda_0=0.05, \lambda_1=0.12$.
All experiments are conducted on NVIDIA RTX A6000 GPUs without any task-specific fine-tuning.

\subsection{Performance Comparison.}

\noindent\textbf{Streaming Video Understanding.} 
~\cref{tab:video_results} and~\cref{tab:results} report results on StreamingBench \cite{lin2024streamingbenchassessinggapmllms} and OVO-Bench \cite{li2025ovobenchfarvideollmsrealworld}.
Across both benchmarks, \textit{all D-HSM variants consistently outperform other methods}.
On StreamingBench, D-HSM raises the overall score from 73.7 to 84.7 using Qwen2.5-VL, and further reaches 85.4 with Qwen3-VL.
D-HSM surpasses the strongest proprietary model, Gemini 1.5 Pro, and the strongest open-source online model, Streamforest, by 9.7 and 8.1 points, respectively.
The improvement suggests that the gains are not tied to a particular backbone, but are closely related to how D-HSM represents and retrieves long-range history.

On OVO-Bench, the improvement is not limited to a single question type: D-HSM achieves competitive results across Real-Time Visual Perception, Backward Tracing, and Forward Active Responding tasks.
This indicates that D-HSM can dynamically adapt its evidence use, relying on the recent visual window for current-state questions while retrieving historical evidence for questions that require earlier moments.

\noindent\textbf{Offline Long Video Understanding.}
~\cref{tab:video_benchmarks_reorg} evaluates D-HSM on offline long-video understanding benchmarks, including LongVideoBench \cite{wu2024longvideobenchbenchmarklongcontextinterleaved}, MLVU \cite{zhou2025mlvubenchmarkingmultitasklong}, and VideoMME \cite{fu2025videommefirstevercomprehensiveevaluation}.
%
%
With Qwen2.5-VL as the backbone, D-HSM achieves 60.7 on LongVideoBench, 67.3 on MLVU, and 63.9 on VideoMME, outperforming the Qwen2.5-VL baseline and several open-source long-video models.
%
%
%
This shows that, beyond the streaming setting, D-HSM can serve as an effective compact representation for long-range evidence in offline long-video understanding.

\begin{table}[t!]
\centering
\normalsize
\caption{\textbf{Evaluation results on offline long-video understanding benchmarks.}
We evaluate D-HSM on LongVideoBench \cite{wu2024longvideobenchbenchmarklongcontextinterleaved}, MLVU \cite{zhou2025mlvubenchmarkingmultitasklong}, and VideoMME \cite{fu2025videommefirstevercomprehensiveevaluation} against different methods. D-HSM results are reported with Qwen2.5-VL using 4 recent frames. 
}
\resizebox{1.0\columnwidth}{!}{
\begin{tabular}{l c | c c c}
\toprule
\textbf{Model} & \textbf{Size} & \textbf{LongVideoBench} & \textbf{MLVU} & \textbf{VideoMME} \\
 & & (Val) & (M-Avg) & (w/o subs) \\
\cmidrule(lr){1-5}
\multicolumn{2}{l}{Video Duration (Avg.)} & 473s & 651s & 1010s \\
\multicolumn{2}{l}{Video Length} & 8s -- 60min & 3min -- 120min & 1min -- 60min \\
\midrule

\multicolumn{5}{c}{\textsc{Proprietary Models}} \\

GPT-4o          & --    & 66.7 & 64.6 & 71.9 \\
Gemini 1.5 Pro  & --    & 64.0 & --   & 75.0 \\
\midrule

\multicolumn{5}{c}{\textsc{Open-Source Offline Models}} \\
LongVA & 7B  & -- &56.3 &52.6 \\
Kangaroo       & 8B  & 54.8 & 61.0 & 56.0 \\
LongVU         & 7B   & --   & 65.4 & 60.6 \\
Apollo         & 7B   & 58.5 & 70.9 & 61.3 \\

SF-LLaVA-1.5   & 7B   & 62.5 & 71.5 & 63.9 \\
InternVL2.5    & 8B   & 60.0 & 68.9 & 64.2 \\
NVILA          & 8B   & 57.7 & 70.1 & 64.2 \\
VideoLLaMA3    & 7B   & 59.8 & 73.0 & 66.2 \\
\midrule
\multicolumn{5}{c}{\textsc{Open-Source Online Models}} \\
Dispider       & 7B   & -- & 61.7 & 57.2 \\
Streamforest   & 7B   & -- & \textbf{69.6} & 61.9 \\
TimeChatOnline & 7B   &  \underline{57.7}  & 65.4 & \underline{62.5}\\

\rowcolor{cyan!7}\textbf{D-HSM} & 7B & \textbf{60.7}  & \underline{67.3} &  \textbf{63.9}  \\ 

\bottomrule
\end{tabular}
}

\label{tab:video_benchmarks_reorg}
\end{table}

\subsection{Ablation Study}

\noindent\textbf{Effect of Memory Sources.}
\begin{table}[t]
\centering
\small
\caption{\textbf{Ablation of memory sources on OVO-Bench.}
We compare HSM only, recent frames only, and the full D-HSM.
All results use Qwen2.5-VL-7B; the recent visual window contains four frames when enabled.
}
\resizebox{0.8\columnwidth}{!}{
\begin{tabular}{lcc}
\toprule
 & \textbf{Real-Time} & \textbf{Backward} \\
\midrule
HSM only                     & 44.79 & 58.02 \\
Recent frames only           & 78.98 & 55.66 \\
\rowcolor{cyan!7} Full D-HSM & \textbf{78.98} & \textbf{62.82} \\
\bottomrule
\end{tabular}
}
\label{tab:ablation_hub_recent_ovo}
\end{table}
~\cref{tab:ablation_hub_recent_ovo} compares the two evidence sources in D-HSM.
Recent frames preserve real-time visual details, whereas HSM provides stronger historical evidence.
Combining them yields the best backward score without sacrificing real-time accuracy, showing that recent perception and long-range memory are necessary.

\noindent\textbf{Effect of Memory Organization.}
\begin{table}[t]
\centering
\small
\setlength{\tabcolsep}{3.5pt}
\caption{\textbf{Ablation of memory organization on OVO-Bench.}
We progressively replace flat captions with typed and entity-centered memory, and remove individual D-HSM components.
All results use Qwen2.5-VL-7B with four recent frames.
}
\resizebox{\columnwidth}{!}{
\begin{tabular}{@{}lcc@{}}
\toprule
 & \textbf{Overall} & \textbf{Backward} \\
\midrule
Flat captions (chronological)                & 58.97 & 53.91 \\
Typed observations w/o entity organization  & 62.99 & 60.73 \\
Entity-centered memory only                  & 63.51 & 60.94 \\
\rowcolor{cyan!7} Full D-HSM               & \textbf{66.80} & \textbf{62.82} \\
\quad w/o co-occurrence edges              & 65.33 & 60.46 \\
\quad w/o next-action chains               & 65.48 & 60.69 \\
\quad w/o spoke merging                    & 66.28 & 61.56 \\
\bottomrule
\end{tabular}
}
\label{tab:ablation_memory_organization}
\end{table}
~\cref{tab:ablation_memory_organization} ablates how historical observations are organized.
Typed observations improve the quality over flat chronological captions, and entity-centered organization provides better results. 
Besides, the D-HSM organization provides further gains, with each component making a meaningful contribution. 
Removing co-occurrence edges, next-action chains, or spoke merging consistently reduces performance, confirming that these components provide complementary temporal and relational evidence.

\noindent\textbf{Effect of Dynamic Retrieval.}
\begin{table}[t]
\centering
\small
\caption{
\textbf{Ablation on dynamic retrieval.}
We compare fixed top-$12$ retrieval with the proposed dynamic retrieval strategy, whose maximum retrieval budget is also set to 12 (Qwen2.5-VL-7B, 4 recent frames).
}
\resizebox{1.0\columnwidth}{!}{

\begin{tabular}{lcc}
\toprule
\textbf{Dataset} &
\textbf{Fixed Top-$k$} &
\textbf{Dynamic} \\
\midrule
OVO-Bench             &  64.71   & \textbf{66.84} \\
StreamingBench  & 79.47 & \textbf{82.45} \\
\bottomrule
\end{tabular}
}

\label{tab:ablation_topk_dynamic}

\end{table}
We compare the proposed dynamic retrieval strategy with a fixed top-$K$ strategy that always retrieves the maximum number 12 of memory entries.
As shown in ~\cref{tab:ablation_topk_dynamic}, dynamic retrieval improves performance on both benchmarks, indicating that adapting retrieval to different question types is beneficial.
%
%

\begin{figure}[t]
    \centering
    \includegraphics[width=\columnwidth]{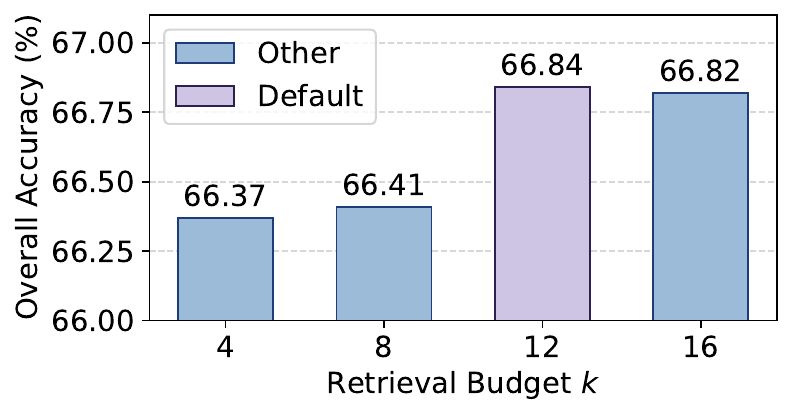}
    \caption{\textbf{Ablation on the maximum retrieval budget $K$ on OVO-Bench} (Qwen2.5-VL-7B, 4 recent frames).}
    \label{fig:topk}
\end{figure}

\noindent\textbf{Effect of Retrieval Budget.}
We vary the maximum retrieval budget $K$ to study how much memory evidence should be made available during retrieval.
As shown in \cref{fig:topk}, performance improves as $K$ increases from 4 to 12, suggesting that a larger budget helps recover supporting long-range evidence.
However, further increasing $K$ to 16 does not bring additional gains and slightly reduces performance.
This suggests that retrieving more memory entries is not always beneficial, since weakly related evidence may introduce unnecessary tokens and distract the model.
The result further supports the need for adapting the amount of retrieved evidence instead of always using a larger fixed budget.

\begin{figure}[t]
    \centering
    \includegraphics[width=\columnwidth]{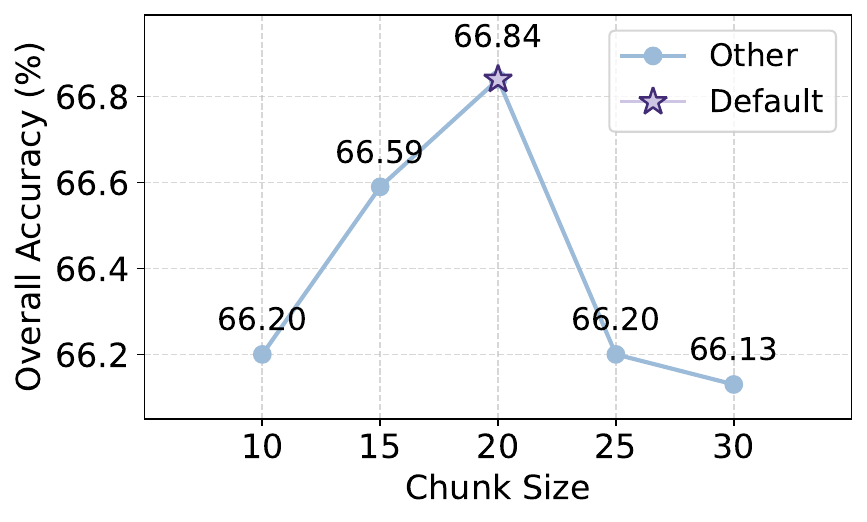}
    \caption{\textbf{Ablation on the historical observation budget on OVO-Bench} (Qwen2.5-VL-7B, 4 recent frames).}
    \label{fig:chunk}
\end{figure}

\noindent\textbf{Effect of Historical Observation Budget.}
We vary the historical observation budget by changing the number of chunks used to construct memory.
As shown in \cref{fig:chunk}, performance first improves as the budget increases, reaching the top with 20 chunks.
This indicates that a larger historical budget can provide broader temporal coverage and preserve more useful long-range evidence.
However, further increasing the budget to 25 or 30 chunks reduces performance.
We attribute this drop to redundant or weakly relevant historical observations, which can introduce noise into memory and make retrieval less focused.
These results suggest that D-HSM benefits from a moderate historical budget that balances coverage and compactness.

\section{Analysis}
\label{sec:analysis}

\subsection{Inference Efficiency Analysis}
\begin{table}[t]
\centering
\small
\setlength{\tabcolsep}{4pt}
\caption{\textbf{End-to-end efficiency of D-HSM.}
Average wall-clock time over 50 StreamingBench samples on a single A100.
Streaming costs are measured per selected chunk or memory rotation, whereas question-time costs are measured per question.
}
\resizebox{\columnwidth}{!}{
\begin{tabular}{lc}
\toprule
\textbf{Stage} & \textbf{Average time} \\
\midrule
Observation generation & 1.55 s \\
Memory update          & 49 ms \\
Removed-chunk handling          & 0.6 ms \\
\midrule
Retrieval            & 11 ms \\
Answer generation          & 0.74 s \\
\bottomrule
\end{tabular}
}
\label{tab:efficiency}
\end{table}
We measure the end-to-end cost of D-HSM on 50 StreamingBench samples using a single Nvidia A100 GPU.
\cref{tab:efficiency} separates background streaming ingestion from the latency after a question arrives.
Observation generation dominates ingestion at 1.55 seconds per selected chunk, whereas the structural memory update takes 49 ms and removed-chunk handling takes only 0.6 ms per rotation.
Video ingestion and memory processing are streamed, so subsequent chunks can arrive while the current selected chunk is being processed. Because the processing is much faster than it accumulates, preventing any long-term backlog.
At question time, retrieval takes 11 ms and answer generation takes 0.74 seconds, giving a question-path latency of approximately 0.75 seconds.
Thus, the hub-and-spoke memory operations add little overhead; most computation comes from the frozen VLM used for observation and answer generation.

\subsection{Dynamic Cutoff Analysis}
\begin{figure}[t]
    \centering
    \includegraphics[width=\linewidth]{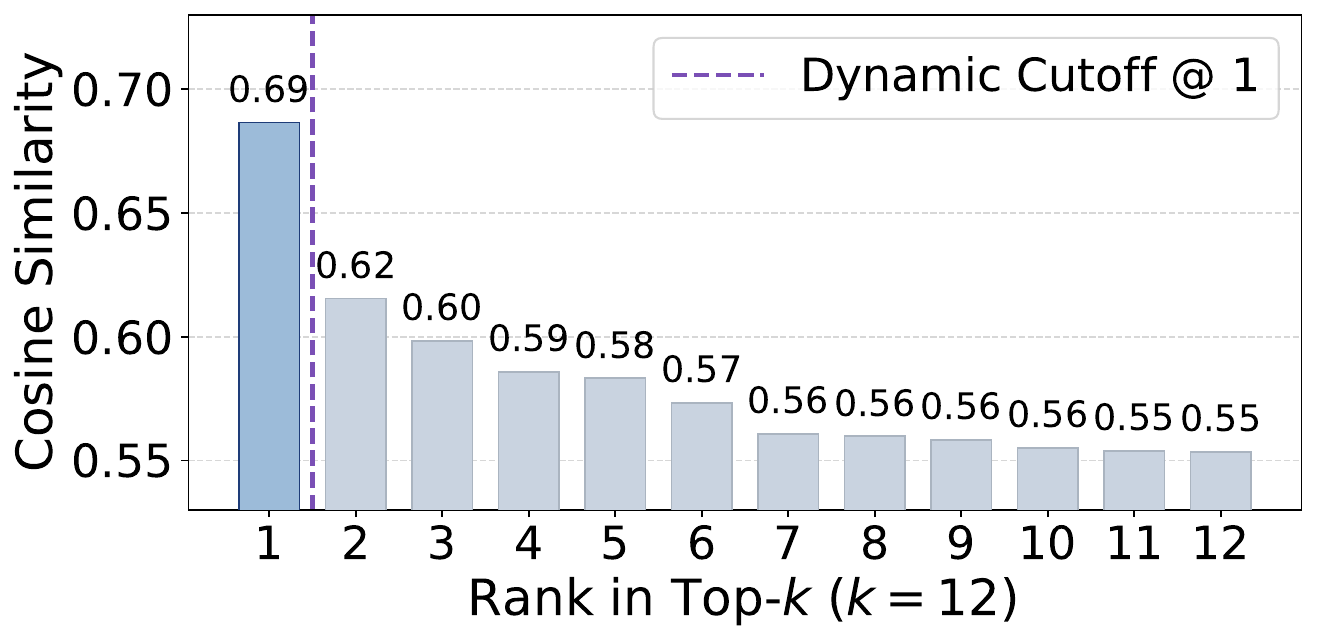}
    \caption{\textbf{Dynamic cutoff analysis for the question ``What does the person do after chopping the onion?''.}
The similarity scores quickly flatten after the high-similarity prefix.}
    \label{fig:similarity}
\end{figure}
We analyze the top-12 question--memory similarities to explain the behavior of dynamic retrieval.
As shown in \cref{fig:similarity}, a short high-similarity prefix is followed by a much flatter score distribution.
D-HSM cuts off retrieval at this gap instead of always retaining all 12 candidates, thereby filtering weakly related entries before hub-and-spoke expansion.
This keeps the evidence compact, reduces distraction from marginal historical context, and lowers the average number of retrieved tokens by about 51\% on OVO-Bench.

\subsection{Memory Quality Analysis}
\begin{table}[t]
\centering
\caption{\textbf{Quality of the constructed memory.}
Human evaluation on 30 StreamingBench videos using Qwen2.5-VL-7B.
}
\resizebox{0.92\columnwidth}{!}{
\begin{tabular}{lcc}
\toprule
\textbf{Metric} & \textbf{Score} & \textbf{Samples} \\
\midrule
Entity-linking pair accuracy $\uparrow$ & 77.0\% & 100 pairs \\
Hub purity  $\uparrow$                 & 84.4\% & 90 hubs \\
Duplicate-hub rate $\downarrow$ & 8.8\% & 170 hubs \\
Spoke-attachment precision $\uparrow$  & 87.1\% & 96 spokes \\
\bottomrule
\end{tabular}
}
\label{tab:memory_quality}
\end{table}
To evaluate the intermediate memory independently, we conduct a comprehensive quantitative evaluation of D-HSM’s key components using human annotations. We randomly sampled 30 videos from StreamingBench and manually annotated by human: (i) 100 pairs of temporally distinct entity mentions that D-HSM merged into the same hub, (ii) 170 entity hubs, and (iii) 96 spoke-to-hub attachments. The 100 mentioned pairs span 90 distinct hubs. 
As shown in \cref{tab:memory_quality}, the accuracy are computed by comparing D-HSM outputs produced with Qwen2.5-VL against human judgments as ground truth. The results identify that the structured memory is reasonably reliable.

\subsection{Failure Cases Analysis}
We manually assign 100 incorrect predictions to mutually exclusive primary causes.
The complete breakdown is reported in \cref{tab:failure_analysis}. The largest source is answer synthesis by the frozen VLM (31\%), followed by observation generation (23\%) and retrieval (19\%).
It shows that in D-HSM, the main bottleneck is evidence selection rather than graph expansion. 
Entity linking accounts for another 10\% of failures, consistent with the memory quality evaluation above.
We find that most of them are predictive questions, because such questions often contain weak entity-specific cues.

\section{Conclusion}
We presented D-HSM, a training-free framework that balances long-range memory and immediate perception for streaming video understanding.
D-HSM stores distant history as entity-centered hub-and-spoke textual memory while keeping recent frames for fine-grained perception.
Its dynamic retrieval selects and expands question-relevant evidence on demand, reducing unnecessary historical context.
Experiments on streaming and offline long-video benchmarks show that D-HSM outperforms existing online and offline baselines.
These results highlight the value of reconstructing the past through structured, question-aware memory.

\clearpage

\section{Limitations}
\label{sec:limitations}
D-HSM relies on frozen VLMs to generate structured textual observations, so errors in entity recognition, OCR, or action descriptions may propagate into memory.
Since long-range history is stored as text, fine-grained details from distant moments may also be lost.
Future work can explore more robust entity linking.

\section{Ethical Considerations}
Our work uses publicly available datasets and open-source models. Potential risks include hallucinated or biased model outputs and possible misuse of streaming video understanding systems in sensitive applications. However, our work does not collect private user data or involve human subjects. We encourage responsible research and deployment practices. 
All datasets and open-source models used in this work are publicly available and are used in accordance with their respective licenses and terms of use.
\section*{Acknowledgments} This research is supported in part by grants from ONR
N00014-21-1-2431, NSF CMMI 2402438, 

{
\small
\bibliography{reference}
}


\appendix
\renewcommand{\thefigure}{\thesection\arabic{figure}}
\renewcommand{\thetable}{\thesection\arabic{table}}
\renewcommand{\theequation}{\thesection\arabic{equation}}

\makeatletter
\@addtoreset{figure}{section}
\@addtoreset{table}{section}
\@addtoreset{equation}{section}
\makeatother

\clearpage

\section*{Appendix}

\section{D-HSM Algorithm}
This section provides the detailed algorithmic procedure of D-HSM.
Algorithm~\ref{alg:dhsm_overview} summarizes the overall streaming inference process.
At each streaming step, D-HSM selects an active set of historical chunks, updates the hub-and-spoke memory with newly selected chunks, and answers a question when it arrives by combining retrieved textual evidence with the recent visual window.
Algorithm~\ref{alg:dhsm_update} details the memory update procedure, including entity-hub creation, spoke insertion and merging, co-occurrence edge construction, and temporal action linking.
Algorithm~\ref{alg:dhsm_retrieval} describes the question-adaptive retrieval procedure, where D-HSM first skips memory retrieval for explicitly current-state questions and otherwise applies dynamic cutoff over the question-memory similarity distribution.

\noindent\textbf{Algorithm~\ref{alg:dhsm_update} notation.}
$\mathcal{H}(M)$ and $\mathcal{P}(M)$ are the hub and spoke sets;
$\mathcal{E}^{\mathrm{co}}(M)$ and $\mathcal{E}^{\mathrm{temp}}(M)$ are the
co-occurrence and temporal-link sets, and $\mathcal{X}(M)$ is their union.
$\mathrm{supp}(x)$ is the set of chunks supporting memory element $x$, and
$\mathcal{I}(c)$ is the chunk-to-memory inverted index.
$n_x$, $T_x$, $t(c)$, and $a(p)$ denote the occurrence count, timestamp set,
chunk timestamp, and spoke-to-hub attachment, respectively.
$\phi$ is the text encoder, $\phi_{\mathrm{obs}}$ is the chunk-observation
extractor, and $\mathcal{N}_j$ and $\mathcal{P}_j$ are the entity and spoke
records extracted from chunk $h_j$.
$\operatorname{id}$ and $\operatorname{IDs}$ denote an entity identity and
the identities referenced by a spoke.
$\operatorname{NN}$ returns the nearest hub and its similarity;
$\operatorname{NewHub}$ and $\operatorname{MergeOrInsert}$ are insertion
operators. $\Psi_{\mathrm{rel}}$ recomputes affected relations, while
$\Psi_{\mathrm{co}}$ and $\Psi_{\mathrm{temp}}$ add co-occurrence and
temporal links.

\section{Structured Observation Template}

\label{app:observation-template}
D-HSM does not store free-form captions as its historical memory.  Each
selected historical segment is a fixed-duration temporal chunk, optionally
chosen by uniform subsampling under the memory budget.  For each selected
chunk, the frozen VLM is prompted to emit a fixed-schema textual observation.
This observation acts as the segment summary used to update the
hub-and-spoke memory.

The template contains six fields:
\textsc{Objects}, \textsc{People}, \textsc{Actions}, \textsc{OCR Text},
\textsc{Spatial}, and \textsc{Event}.  In the implementation, the prompt uses
the header \texttt{TEXT}; we refer to this field as \textsc{OCR Text} in the
paper because it is intended to record readable on-screen text rather than
arbitrary generated prose.  The VLM is instructed to output \texttt{NONE} when
a field has no visible evidence.

\begin{algorithm}[t]
\caption{Overall D-HSM Streaming Inference}
\label{alg:dhsm_overview}
\begin{algorithmic}[1]
\Require Historical chunks $\{h_t\}_{t=1}^{T}$; recent visual window $r_t$; question $q_t$
\Require History budget $B$; retrieval budget $K$; threshold $\theta$; gap parameters $\lambda_0,\lambda_1$
\Ensure Answer $a_t$ when question $q_t$ arrives
\State Initialize memory $M_0 \gets \emptyset$ and active set $\mathcal{S}_0 \gets \emptyset$
\For{$t=1,\ldots,T$}
    \State $\mathcal{H}_t \gets \{h_1,\ldots,h_t\}$
    \State $\mathcal{S}_t \gets \mathcal{H}_t$ if $t \le B$, otherwise $\Pi_B(\mathcal{H}_t)$
    \State $\Delta_t^+ \gets \mathcal{S}_t \setminus \mathcal{S}_{t-1}$
    \State $\Delta_t^- \gets \mathcal{S}_{t-1} \setminus \mathcal{S}_t$
    \State $M_t \gets \Call{UpdateMemory}{M_{t-1},\Delta_t^+,\Delta_t^-}$ \Comment{Alg.~\ref{alg:dhsm_update}}
    \If{question $q_t$ arrives}
        \State $C_t \gets \Call{Retrieve}{M_t,q_t,K,\theta,\lambda_0,\lambda_1}$ \Comment{Alg.~\ref{alg:dhsm_retrieval}}
        \State $a_t \gets f_{\mathrm{VLM}}(C_t,r_t,q_t)$
    \EndIf
\EndFor
\end{algorithmic}
\end{algorithm}
\begin{algorithm}[t]
\caption{Hub-and-Spoke Memory Update}
\label{alg:dhsm_update}
\begin{algorithmic}[1]
\Require Previous memory $M_{t-1}$; added chunks $\Delta_t^+$;
removed chunks $\Delta_t^-$; entity-linking threshold $\tau_{\mathrm{link}}$
\Ensure Updated memory $M_t$
\Statex $\mathcal{X}(M)=\mathcal{H}(M)\cup\mathcal{P}(M)\cup
\mathcal{E}^{\mathrm{co}}(M)\cup\mathcal{E}^{\mathrm{temp}}(M)$
\Statex $\mathcal{I}(c)=\{x\in\mathcal{X}(M):c\in\mathrm{supp}(x)\}$
\State $M_t \gets M_{t-1}$
\ForAll{$c\in\Delta_t^-,\;x\in\mathcal{I}(c)$}
    \State $\mathrm{supp}(x)\gets\mathrm{supp}(x)\setminus\{c\}$
\EndFor
\State $M_t\gets M_t\setminus
\{x\in\mathcal{X}(M_t):\mathrm{supp}(x)=\emptyset\}$
\State $(n_x,T_x)\gets
\bigl(|\mathrm{supp}(x)|,\{t(c):c\in\mathrm{supp}(x)\}\bigr),
\ \forall x\in\mathcal{X}(M_t)$
\State $a(p)\gets\arg\max_{v\in\mathcal{H}(M_t)}
\cos\!\left(\phi(p),\phi(v)\right),
\ \forall p:a(p)\notin\mathcal{H}(M_t)$
\State $(\mathcal{E}^{\mathrm{co}},\mathcal{E}^{\mathrm{temp}})
\gets\Psi_{\mathrm{rel}}(M_t)$
\ForAll{$h_j\in\Delta_t^+$}
    \State $z_j\gets\phi_{\mathrm{obs}}(h_j)$
    \State $\mathcal{N}_j\gets
    z_j[\textsc{People}\cup\textsc{Objects}]$
    \State $\mathcal{P}_j\gets
    z_j[\textsc{Actions}\cup\textsc{OCR-Text}\cup
    \textsc{Spatial}\cup\textsc{Event}]$
    \ForAll{$u\in\mathcal{N}_j$}
        \State $(v^\star,s^\star)\gets
        \operatorname{NN}\!\left(\phi(u),\mathcal{H}(M_t)\right)$
        \If{$s^\star\geq\tau_{\mathrm{link}}$}
            \State $\operatorname{id}(u)\gets\operatorname{id}(v^\star)$
        \Else
            \State $v^\star\gets\operatorname{NewHub}(u)$
            \State $M_t\gets M_t\cup\{v^\star\}$
        \EndIf
        \State $\mathrm{supp}(v^\star)\gets
        \mathrm{supp}(v^\star)\cup\{h_j\}$
    \EndFor
    \ForAll{$p\in\mathcal{P}_j$}
        \If{$\operatorname{IDs}(p)\neq\emptyset$}
            \State $a(p)\gets
            \{v:\operatorname{id}(v)\in\operatorname{IDs}(p)\}$
        \Else
            \State $a(p)\gets\arg\max_{v\in\mathcal{H}(M_t)}
            \cos\!\left(\phi(p),\phi(v)\right)$
        \EndIf
        \State $p^\star\gets\operatorname{MergeOrInsert}(p,M_t)$
        \State $\mathrm{supp}(p^\star)\gets
        \mathrm{supp}(p^\star)\cup\{h_j\}$
    \EndFor
    \State $\mathcal{E}^{\mathrm{co}}\gets
    \mathcal{E}^{\mathrm{co}}\cup\Psi_{\mathrm{co}}(\mathcal{N}_j,h_j)$
    \State $\mathcal{E}^{\mathrm{temp}}\gets
    \mathcal{E}^{\mathrm{temp}}\cup\Psi_{\mathrm{temp}}(\mathcal{P}_j,h_j)$
    \State $\mathcal{I}(h_j)\gets
    \{x\in\mathcal{X}(M_t):h_j\in\mathrm{supp}(x)\}$
\EndFor
\State \Return $M_t$
\end{algorithmic}
\end{algorithm}
\begin{algorithm}[t]
\caption{Dynamic Hub-and-Spoke Retrieval}
\label{alg:dhsm_retrieval}
\begin{algorithmic}[1]
\Require Memory $M_t$; question $q_t$; retrieval budget $K$;
base threshold $\theta$; gap parameters $\lambda_0,\lambda_1$
\Ensure Linearized textual context $C_t$
\Statex $\mathcal{X}_t\!=\!\mathcal{H}(M_t)\cup\mathcal{P}(M_t)$:
memory entries; $\phi$: text encoder
\Statex $\chi_{\mathrm{cur}},\chi_{\mathrm{time}}$:
current-state and time-aware query indicators
\Statex $\operatorname{TopK}_K$: top-$K$ by score;
$\operatorname{sort}_{\downarrow}$: descending score sort
\Statex $\Gamma_{\mathrm{sp}},\Gamma_{\mathrm{co}},
\Gamma_{\mathrm{temp}}$: spoke, co-occurrence, and temporal-link expansion
\Statex $\mathcal{T}$: timestamp augmentation;
$\operatorname{Lin}$: textual linearization
\If{$\chi_{\mathrm{cur}}(q_t)=1$}
    \State \Return $C_t\gets\emptyset$
\EndIf
\State $s(e)\gets\cos\!\left(\phi(q_t),\phi(e)\right),
\ \forall e\in\mathcal{X}_t$
\State $\mathcal{A}_t\gets
\operatorname{TopK}_K\{e\in\mathcal{X}_t:s(e)\geq\theta\}$
\State $\{(e_i,s_i)\}_{i=1}^{n}\gets
\operatorname{sort}_{\downarrow}
\{(e,s(e)):e\in\mathcal{A}_t\}$
\If{$n<2$}
    \State $k_t\gets n$
\Else
    \State $\Delta_i\gets s_i-s_{i+1},
    \quad i\in\{1,\ldots,n-1\}$
    \State $i^\star\gets\arg\max_i\Delta_i,
    \quad \Delta^\star\gets\Delta_{i^\star}$
    \State $\gamma_t\gets
    \max\!\left(\lambda_0,\lambda_1[s_1-\theta]_+\right)$
    \If{$\Delta^\star\geq\gamma_t\ \land    \Delta^\star\geq2\,\operatorname{median}_i(\Delta_i)$}
        \State $k_t\gets i^\star$
    \Else
        \State $k_t\gets n$
    \EndIf
\EndIf
\State $\mathcal{E}_t\gets\{e_i\}_{i=1}^{k_t}$
\State $\widehat{\mathcal{E}}_t\gets
\mathcal{E}_t\cup\Gamma_{\mathrm{sp}}(\mathcal{E}_t)$
\State $\widehat{\mathcal{E}}_t\gets
\widehat{\mathcal{E}}_t\cup
\Gamma_{\mathrm{co}}(\widehat{\mathcal{E}}_t)\cup
\Gamma_{\mathrm{temp}}(\widehat{\mathcal{E}}_t)$
\If{$\chi_{\mathrm{time}}(q_t)=1$}
    \State $\widehat{\mathcal{E}}_t\gets
    \widehat{\mathcal{E}}_t\cup
    \mathcal{T}(\widehat{\mathcal{E}}_t)$
\EndIf
\State $C_t\gets\operatorname{RenderText}(\widehat{\mathcal{E}}_t)$
\State \Return $C_t$
\end{algorithmic}
\end{algorithm}
\begin{table*}[t]
\centering
\small
\caption{\textbf{Examples of textual evidence blocks rendered from D-HSM memory and provided to the VLM.}
The exact entries depend on the retrieved hubs, spokes, screen-text nodes, and timeline events.}
\begin{tabular}{p{0.22\linewidth} p{0.70\linewidth}}
\toprule
\textbf{Rendered block} & \textbf{Template shown to the VLM} \\
\midrule
Entity-centered profile &
\texttt{[P1 (blue-shirt woman)]}\\
& \texttt{\ \ [00:05] picks up O2 (red mug)}\\
& \texttt{\ \ [00:08] walks toward O3 (wooden table)}\\
\midrule
Screen text &
\texttt{[Screen Text]}\\
& \texttt{\ \ * "EXIT"}\\
& \texttt{\ \ * "12:30"} \\
\midrule
Most recent events &
\texttt{[Most recent events]}\\
& \texttt{\ \ [00:21] The person leaves the room.}\\
& \texttt{\ \ [00:24] The door closes.} \\
\midrule
Timeline &
\texttt{[Timeline]}\\
& \texttt{\ \ [00:05] A woman picks up a red mug near the table.}\\
\midrule
Counting aggregate &
\texttt{[Counting Aggregate --- observed action occurrences across captioned chunks]}\\
& \texttt{\ \ 'place' (e.g. "P1 places O2 on O3"): 3 time(s) at [00:05, 00:08, 00:13]} \\
\bottomrule
\end{tabular}

\label{tab:rendered-context}
\end{table*}

{\small
\begin{verbatim}
OBJECTS: <semicolon-separated visible objects>
PEOPLE: <semicolon-separated visible people>
ACTIONS: <semicolon-separated actions>
TEXT: <readable on-screen text>
SPATIAL: <semicolon-separated spatial relations>
EVENT: <one sentence summarizing the main event>
\end{verbatim}
}

\paragraph{Objects and People.}
The object and person fields provide chunk-level entity mentions used to form entity hubs.  Objects are written with
local identifiers $O_1, O_2, \ldots$ and concise visual attributes such as
color, size, material, or state.  People are written with local identifiers
$P_1, P_2, \ldots$, visible clothing or role cues, and a brief action.
Across segments, D-HSM merges mentions of the same entity into one persistent identity based on visual-description similarity.

\paragraph{Actions.}
The action field records interactions among visible entities, preferably using
the stable identifiers.  For example, an action may state that $P_1$ holds
$O_2$, points toward $O_3$, or talks to $P_2$.  These entries become action
spokes linked to the corresponding entity hubs.

\paragraph{OCR Text.}
The OCR text field records readable text visible in the frame sequence,
including slide text, subtitles, signs, labels, charts, and interface text.
The prompt asks the VLM to copy this text verbatim when readable.  These
entries are stored as screen-text nodes and are retrieved separately for
questions that depend on visual text.

\paragraph{Spatial Relations.}
The spatial field records positional relations between named entities, such as
left/right, above/below, behind, or in front of.  These relations preserve
local visual layout in a textual form.

\paragraph{Event.}
The event field is a single-sentence summary of the most important thing
happening in the segment.  Unlike the entity and relation fields, this field
is intended to provide a compact chronological event trace for temporal
retrieval.

\section{More Ablation Experiments}
\label{app:more-ablation}
\begin{table}[t]
\centering
\caption{\textbf{Ablation on the recent visual window size.}}
\footnotesize 
\resizebox{\columnwidth}{!}{
\begin{tabular}{l cccc} 
\toprule
 & \multicolumn{2}{c}{\textbf{OVO-Bench}} & \multicolumn{2}{c}{\textbf{StreamingBench}} \\
\cmidrule(lr){2-3} \cmidrule(lr){4-5} 
{\textbf{Recent}}& Qwen2.5 & Qwen3 & Qwen2.5 & Qwen3 \\
{\textbf{Frames}}& \scriptsize{(VL-7B)} & \scriptsize{(VL-8B)} & \scriptsize{(VL-7B)} & \scriptsize{(VL-8B)} \\
\midrule
2f & 64.93          & 64.01          & 80.46          & 81.18          \\
4f & \textbf{66.84} & \textbf{65.59} & 82.45          & 83.08          \\
6f & 66.53          & 64.69          & 83.42          & 84.86          \\
8f & 65.55          & 65.09          & \textbf{84.73} & \textbf{85.36} \\
\bottomrule
\end{tabular}
}
\label{tab:ablation_transpose_top_ours}
\end{table}
\begin{table*}[t!]
\centering
\caption{\textbf{Ablation study of different historical processing strategies on OVO-Bench.} HSM means Fixed Hub-and-Spoke Memory, and D-HSM means Dynamic Hub-and-Spoke Memory. The setting of HSM for Top-$k$ is $k=12$. For Ours and HSM Only, the chunk size = 20. Ours in this table is D-HSM with 4 recent frames, and the backbone is Qwen2.5-VL-7B.}
\normalsize
\resizebox{\linewidth}{!}{
\begin{tabular}{l|ccccccc|cccc}
\toprule
\multirow{2}{*}{\textbf{Model}} & \multicolumn{7}{c|}{\textbf{Real-Time}} & \multicolumn{4}{c}{\textbf{Backward}} \\ 

\cmidrule(lr){2-8} \cmidrule(lr){9-12} 
 & OCR & ACR & ATR & STU & FPD & OJR & \textbf{Avg.} & EPM & ASI & HLD & \textbf{Avg.} \\ 

\midrule
HSM Only &  44.30 & 43.12 & 47.41 & 42.13 & 56.44 & 35.33 & 44.79 & 28.62 & 54.05 & 91.40 & 58.02  \\
Recent 4 Frames Only & 93.96 & 70.64 & 86.21 & 66.85 & 75.25 & 80.98 & 78.98 & 54.55 & 60.81 & 51.61 & 55.66 \\
\rowcolor{cyan!7} \textbf{Ours} \textit{Qwen2.5-VL+D-HSM (4f)} & 93.96 & 70.65 & 86.21 & 66.85 & 75.25 & 80.98 & 78.98 
& 52.87 & 63.54 & 72.04 & 62.82   \\
\bottomrule
\end{tabular}
}

\label{tab:hub_recent_ovo_all}
\end{table*}
\begin{table*}[t]
\centering
\normalsize
\caption{
 \textbf{Ablation study of different historical processing strategies on StreamingBench \cite{lin2024streamingbenchassessinggapmllms}}. Our proposed method, Qwen2.5-VL+D-HSM (4f), significantly outperforms both the "HSM Only" and "Recent 4 Frames Only" baselines in overall performance, demonstrating that our Dynamic Hub-and-Spoke Memory (D-HSM) architecture achieves a superior balance between historical context retention and real-time responsiveness.}
\resizebox{\linewidth}{!}{
\begin{tabular}{l|cccccccccc|c}
\toprule
\textbf{Model}  & \textbf{OP} & \textbf{CR} & \textbf{CS} & \textbf{ATP} & \textbf{EU} & \textbf{TR} & \textbf{PR} & \textbf{SU} & \textbf{ACP} & \textbf{CT} & \textbf{Overall} \\
\midrule
HSM Only  & 54.50 & 47.20 & 68.75 & 60.00 & 59.49 & 53.58 & 64.76 & 38.62 & 46.06 & 35.56 & 53.72 \\
Recent 4 Frames Only  & 85.29 & 64.00 & 90.79  & 88.20 & 73.42 & 90.03 & 76.19 & 79.27 & 78.72 & 36.56 & 81.22 \\
\rowcolor{cyan!7}\textbf{Ours} \textit{Qwen2.5-VL+D-HSM (4f)} & {85.29} & {64.00} & {88.82} & {87.87} & {79.11} & {90.03} & 87.62 & 79.67 & {79.59} & {47.78} & {82.45} \\
\bottomrule
\end{tabular}
}

\label{tab:hub_recent_sb_all}
\end{table*}

\noindent\textbf{Effect of Recent Visual Window Size.}
~\cref{tab:ablation_transpose_top_ours} studies the effect of the recent visual window.
Increasing the number of recent frames consistently improves StreamingBench, confirming the importance of fine-grained current perception for real-time streaming questions.
In contrast, OVO-Bench peaks at $4$ frames, suggesting that more recent visual context is not always beneficial when questions also require backward or temporal evidence.
Together, these results show that D-HSM benefits from balancing recent visual perception with retrieved historical memory.

We compare three historical processing strategies: using only the proposed Hub-and-Spoke Memory (HSM), using only recent frames, and the full Qwen2.5-VL+D-HSM (4f) (Ours) framework. \\

\noindent\textbf{Effect of different historical processing strategies.} 
As shown in \cref{tab:hub_recent_ovo_all}, for OVO-Bench, the method that using HSM Only has better performance than the one using Only Recent 4 Frames on Backward Tasks. While the Recent 4 Frames Only baseline achieves strong real-time performance, it suffers from degraded backward reasoning capability. In contrast, our full method significantly improves backward understanding while preserving strong real-time performance, demonstrating that the proposed D-HSM architecture effectively balances long-term historical context retention and real-time responsiveness.
\\

For StreamingBench, our ablation results demonstrate the importance of jointly modeling recent observations and long-term historical memory. Using only the proposed HSM mechanism leads to substantially degraded performance, indicating that relying solely on compressed historical memory is insufficient for fine-grained real-time understanding. In contrast, the ``Recent 4 Frames Only'' baseline achieves strong performance on real-time perception tasks but exhibits limited capability in tasks requiring longer temporal reasoning and contextual consistency. By integrating recent frame observations with the proposed Dynamic Hub-and-Spoke Memory (D-HSM) architecture, Qwen2.5-VL+D-HSM (4f) achieves the best overall performance, improving the overall score from 81.22 to 82.45 while yielding notable gains on temporally demanding tasks such as PR and CT. These results demonstrate that D-HSM effectively balances short-term responsiveness with long-term contextual retention in streaming video understanding.

\section{Detailed Visualization of D-HSM}
\label{app:case-study}

\begin{figure*}[t]
    \centering
    \includegraphics[width=\linewidth]{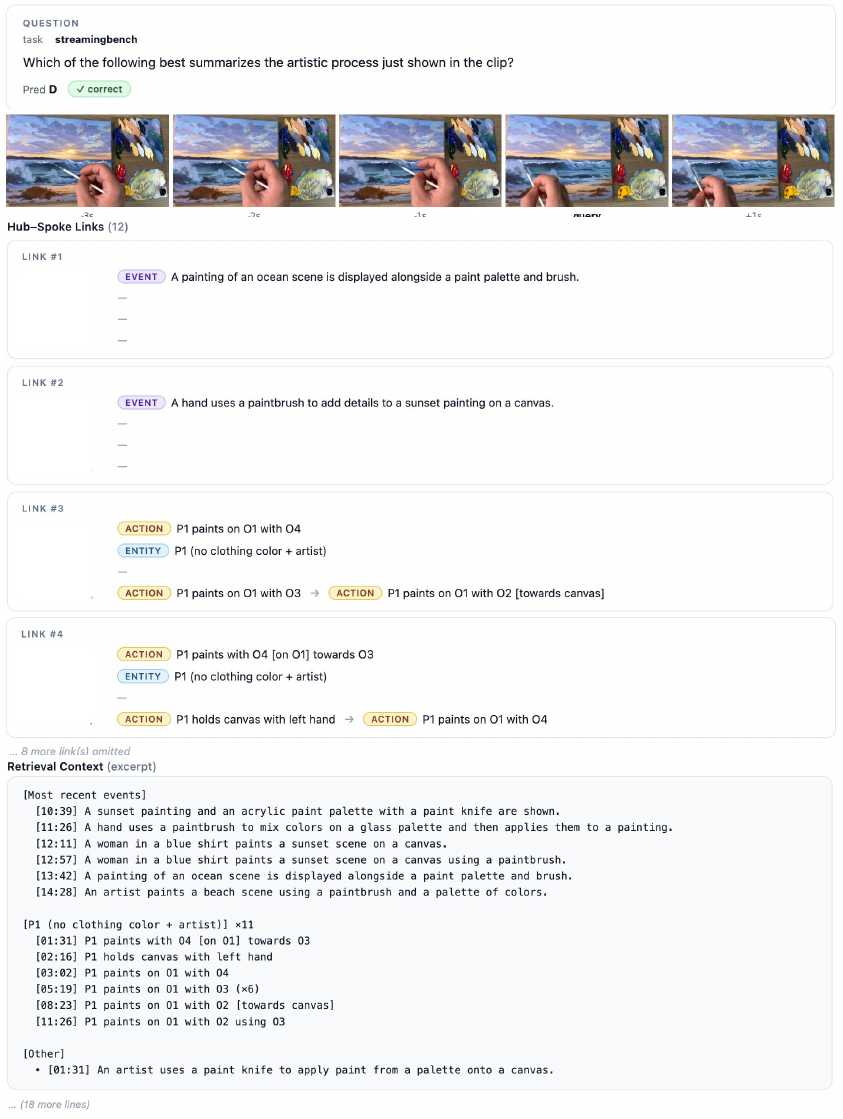}
    \caption{\textbf{Visualization of hub-and-spoke memory details and the integrated retrieval context for the question from StreamingBench.}}
    \label{fig:case1}
\end{figure*}
\begin{figure*}[t]
    \centering
    \includegraphics[width=\linewidth]{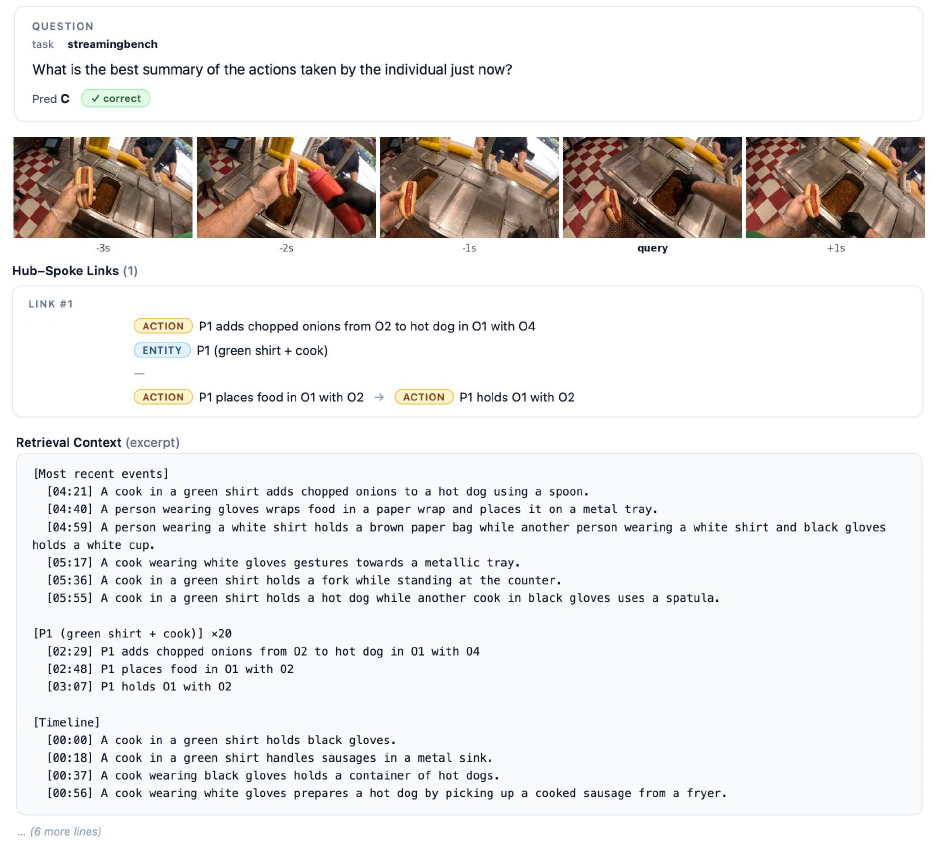}
    \caption{\textbf{Visualization of hub-and-spoke memory details and the integrated retrieval context for the question from StreamingBench.}}
    \label{fig:case2}
\end{figure*}
\begin{figure*}[t]
    \centering
    \includegraphics[width=\linewidth]{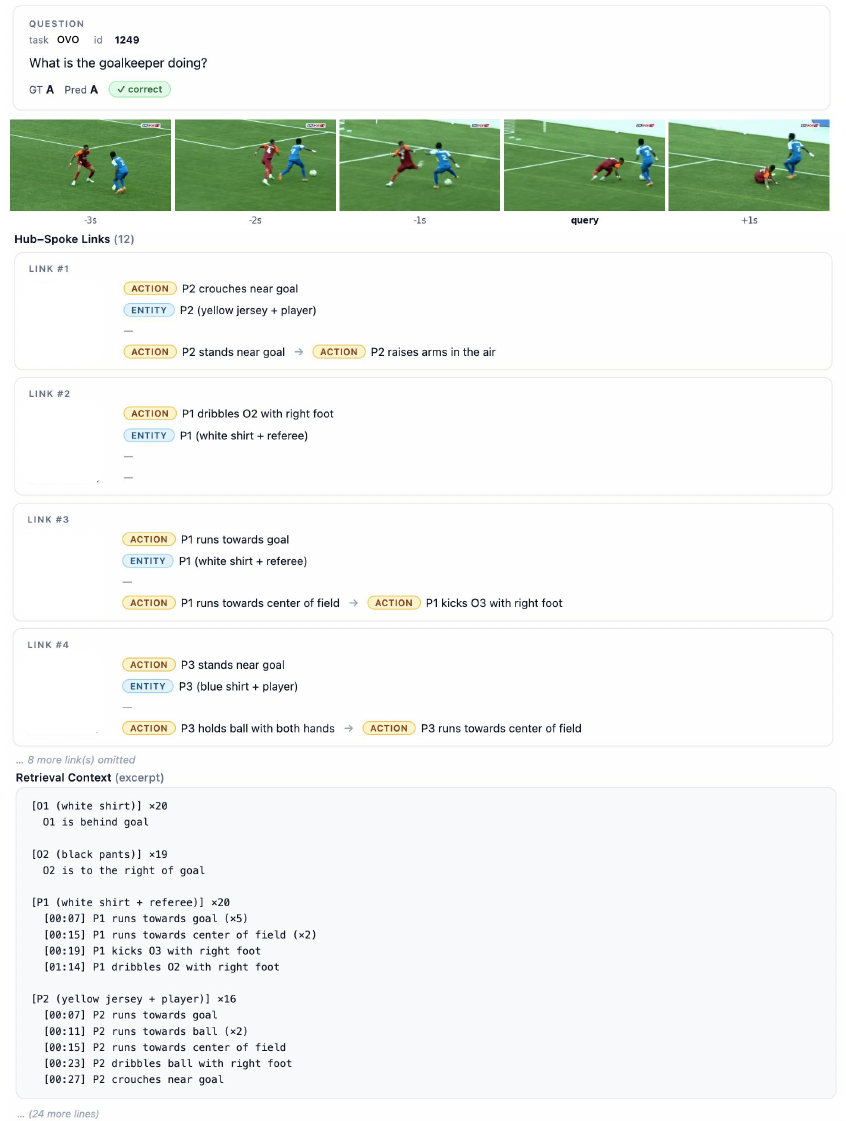}
    \caption{\textbf{Visualization of hub-and-spoke memory details and the integrated retrieval context for the question from OVO-Bench.}}
    \label{fig:case3}
\end{figure*}
\begin{figure*}[t]
    \centering
    \includegraphics[width=\linewidth]{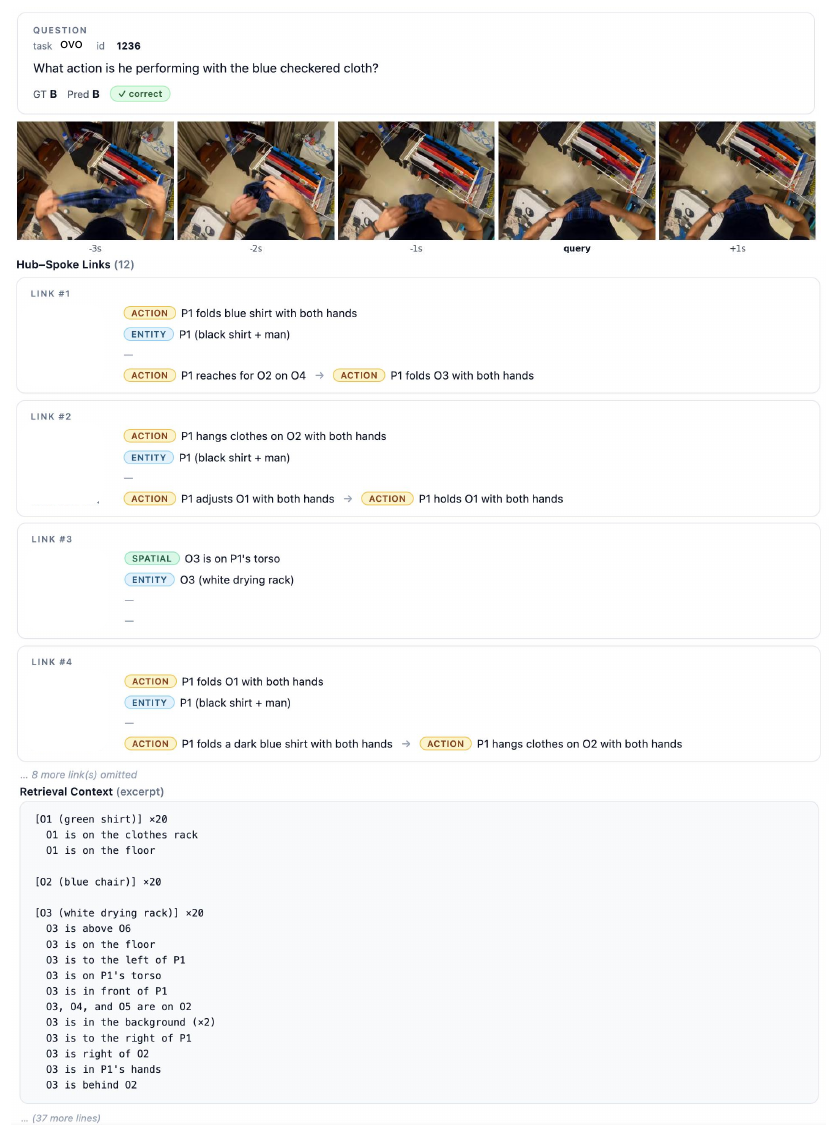}
    \caption{\textbf{Visualization of hub-and-spoke memory details and the integrated retrieval context for the question from OVO-Bench.}}
    \label{fig:case4}
\end{figure*}
\cref{fig:case1,fig:case2} visualize detailed examples of hub-and-spoke memory and integrated retrieval context by D-HSM for StreamingBench questions, while \cref{fig:case3,fig:case4} show examples for OVO-Bench questions.
Given the recent visual window and the question, D-HSM retrieves hub-and-spoke links from memory and renders them into a textual context for the frozen VLM.
The retrieved links include event spokes, entity-centered action spokes, and short temporal action chains.
The final retrieval context corresponds to the rendered evidence format shown in \cref{tab:rendered-context}, but provides a more detailed example with concrete retrieved links and entity-specific evidence.
It illustrates how D-HSM reconstructs question-relevant history instead of replaying all previous chunks.


\section{Use of AI Assistants}
AI assistants were used for language polishing and minor coding assistance. All technical content, experiments, and conclusions were verified by the authors.


\section{Failure Analysis Breakdown}
\label{sec:failure_breakdown}
\begin{table}[t]
\centering
\small
\caption{\textbf{Primary causes of 100 manually analyzed failures.}
Indented rows decompose their parent categories and are not additional errors.
}
\resizebox{0.82\columnwidth}{!}{
\begin{tabular}{lc}
\toprule
\textbf{Error source} & \textbf{Percentage} \\
\midrule
Observation generation       & 23\% \\
\quad Recognition           & 21\% \\
\quad OCR                   & 2\% \\
Entity linking               & 10\% \\
Retrieval                    & 19\% \\
\quad Selection             & 19\% \\
\quad Expansion             & 0\% \\
Answer synthesis             & 31\% \\
Benchmark issues             & 17\% \\
\quad Ground-truth error    & 5\% \\
\quad Multiple valid answers & 12\% \\
\bottomrule
\end{tabular}
}
\label{tab:failure_analysis}
\end{table}
We label each incorrect prediction with one mutually exclusive primary cause.
Observation-generation errors cover incorrect visual recognition or OCR; entity-linking errors merge distinct entities or split one entity across hubs; retrieval errors arise when the required evidence is not selected or expanded; and answer-synthesis errors occur when the frozen VLM produces an incorrect answer despite the available context.
The indented rows in \cref{tab:failure_analysis} decompose their parent categories and therefore are not counted again in the total.

\section{Model Size and Budget}
The model sizes are 7B for Qwen2.5-VL, 8B for Qwen3-VL, and 33M for BGE-Small-EN-v1.5. Regarding inference time, since our method is training-free, there is no training GPU budget. For evaluation on StreamingBench, it takes approximately 1 hour and 20 minutes on 8 A6000 GPUs.

\section{Dataset Statistics}

We summarize the datasets used in our experiments, including StreamingBench \cite{lin2024streamingbenchassessinggapmllms} and OVO-Bench \cite{li2025ovobenchfarvideollmsrealworld}, LongVideoBench \cite{wu2024longvideobenchbenchmarklongcontextinterleaved}, MLVU \cite{zhou2025mlvubenchmarkingmultitasklong}, and and VideoMME \cite{fu2025videommefirstevercomprehensiveevaluation}. We used test only splits for evaluation since our method is training-free.


\end{document}